\documentclass[letterpaper, 10 pt, conference]{ieeeconf}  

\IEEEoverridecommandlockouts                              

\title{\LARGE \bf
False Positive Sampling-based Data Augmentation 
\\for Enhanced 3D Object Detection Accuracy
}

\author{Jiyong Oh$^{1}$, Junhaeng Lee$^{2}$, Woongchan Byun$^{2}$, Minsang Kong$^{2}$ and Sang Hun Lee$^{2, *}$
\thanks{$^{1}$Department of Automotive Engineering, Kookmin University, Seoul, Republic of Korea 
        {\tt\small din2346@kookmin.ac.kr}}%
\thanks{$^{2}$Department of Automotive and IT Convergence, Kookmin University, Seoul, Republic of Korea
        {\tt\small \{junkil564, jans1541, gms0725, shlee\}@kookmin.ac.kr}}%
\thanks{$^{*}$ Corresponding Author}%
\thanks{This work was supported by the National Research Foundation of Korea (NRF) grant funded by the Korea government (MSIT). (No. 2020R1A2C1102767)}%
}

\date{\today}

\usepackage{amsfonts}
\usepackage{amsmath}
\usepackage{amssymb} 
\usepackage{algorithm}
\usepackage{algpseudocode}
\usepackage{graphicx}
\usepackage{multirow}
\usepackage{hyperref}
\usepackage{utfsym}
\usepackage{cite}
\usepackage{longtable}
\usepackage{colortbl}
\definecolor{skyblue}{rgb}{0.88,1,1}

\usepackage{subfig}
\makeatletter
\def\endfigure{\end@float}
\makeatother

\begin{document}

\maketitle

\begin{abstract}

Recent studies have focused on enhancing the performance of 3D object detection models. Among various approaches, ground-truth sampling has been proposed as an augmentation technique to address the challenges posed by limited ground-truth data. However, an inherent issue with ground-truth sampling is its tendency to increase false positives. Therefore, this study aims to overcome the limitations of ground-truth sampling and improve the performance of 3D object detection models by developing a new augmentation technique called false-positive sampling. False-positive sampling involves retraining the model using point clouds that are identified as false positives in the model's predictions. We propose an algorithm that utilizes both ground-truth and false-positive sampling and an algorithm for building the false-positive sample database. Additionally, we analyze the principles behind the performance enhancement due to false-positive sampling. Our experiments demonstrate that models utilizing false-positive sampling show a reduction in false positives and exhibit improved object detection performance. On the KITTI and Waymo Open datasets, models with false-positive sampling surpass the baseline models by a large margin. The code is available at \url{https://github.com/KaAI-KMU/Openpcdet_Sampling}

\end{abstract}
\begin{figure}[ht]
    \centering
    \subfloat[default]{\includegraphics[width=1.0\linewidth, height=0.35\linewidth]{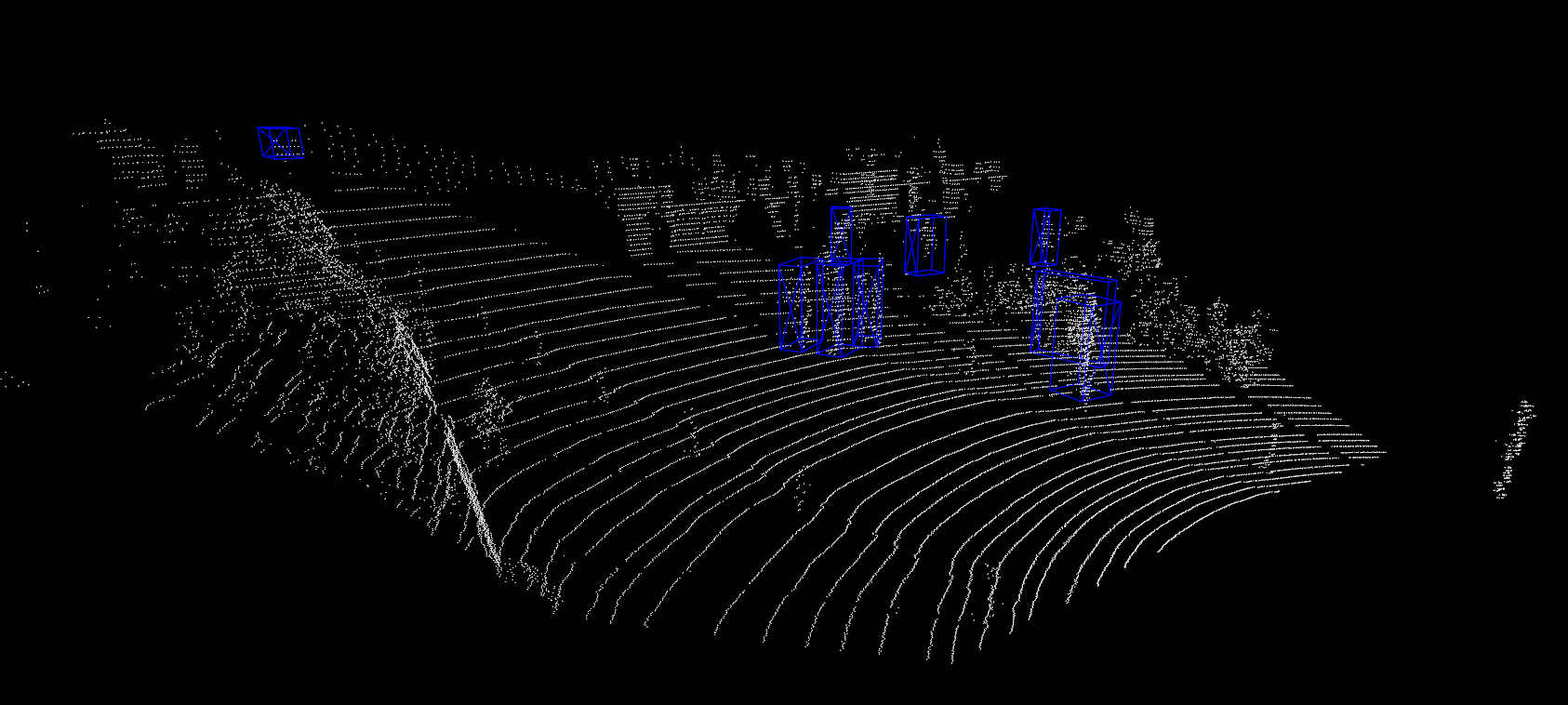}}
    \label{fig:sub1}\\ 
    \subfloat[GT sampling]{\includegraphics[width=1.0\linewidth, height=0.35\linewidth]{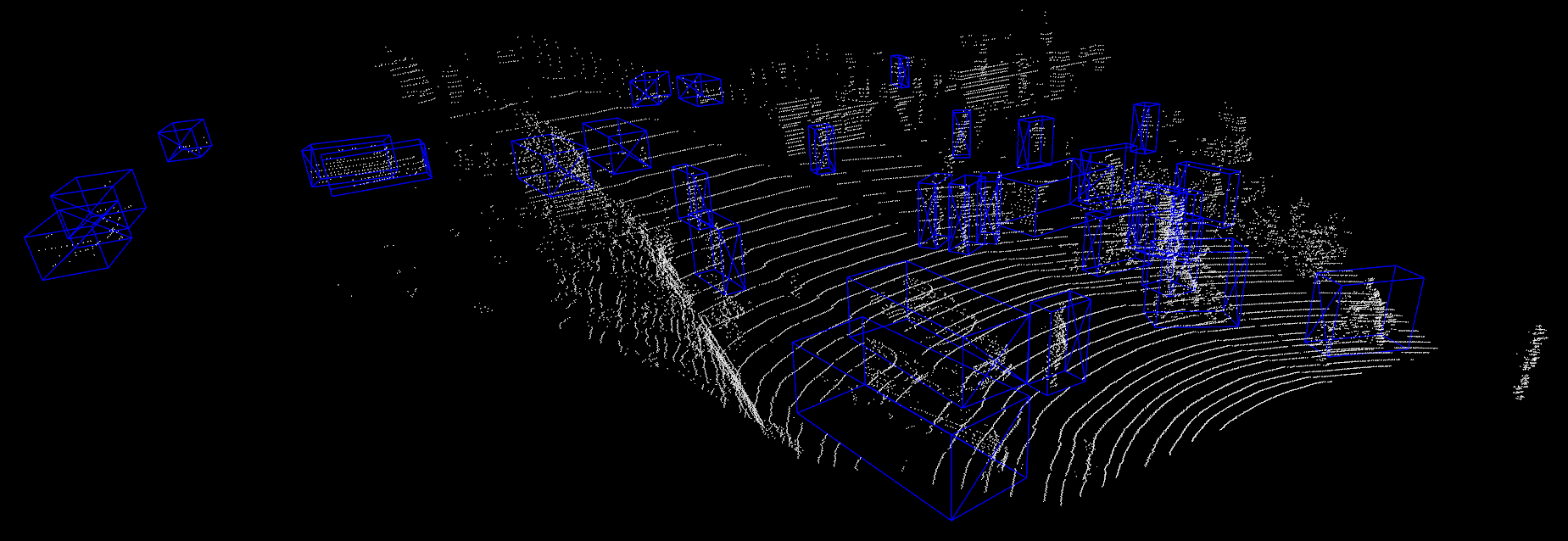}}
    \label{fig:sub2}\\ 
    \subfloat[GT and FP sampling]{\includegraphics[width=1.0\linewidth, height=0.35\linewidth]{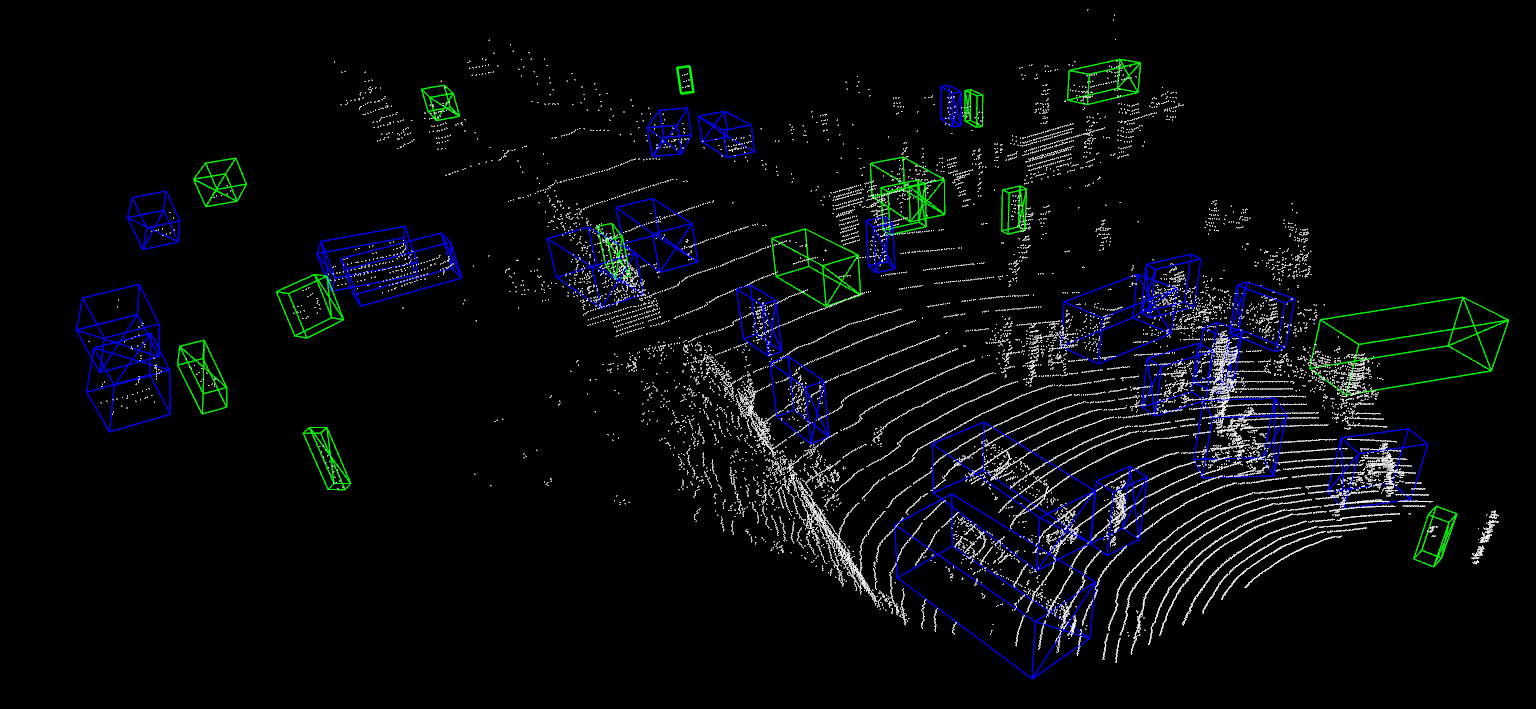}}
    \label{fig:sub3}
    \caption{Comparison of data augmentation methods in 3D object detection: (a) the original point cloud without any data augmentation via GT or FP sampling, (b) data augmentation via GT sampling, and (c) data augmentation via GT and FP sampling. Blue boxes represent ground truths, while green boxes indicate false positives.}
    \label{fig:Visualization}
\end{figure}

\section{INTRODUCTION}
In recent years, significant advancements have been made in 3D object detection, a key area in computer
vision. 3D object detection involves precisely identifying objects in three-dimensional space and estimating their location, size, and orientation. This task is often conducted using point cloud data acquired from LiDAR and is particularly crucial when applied to autonomous driving systems. In this task, detection accuracy is vital to ensure the safety of all agents on the road and facilitate the vehicle's navigation to its intended destination. 

Recent studies have developed various model architectures to efficiently process LiDAR point cloud data,
focusing on highlighting key features in complex 3D road scenes and filtering out noise and irrelevant data. Various data augmentation techniques are applied to train these models. These techniques reflect the
complexity and diversity of real-world environments, make the model learn of more generalized patterns
independent of specific training data, and increase robustness against noise, scale, and orientation.

Ground-truth (GT) sampling \cite{second:}, a data augmentation technique, has been proposed to address the scarcity of labeled data. This method involves getting points corresponding to the ground truth from a GT database and merging them with the current scene for training models. By significantly increasing the ground truth data, GT sampling plays a crucial role in enhancing the generalization ability of models for data with limited or insufficient labels for specific categories. This approach is precious for improving the robustness and effectiveness of object detection models where labeled data is limited.

However, applying GT sampling alone can increase false positives as the model may become over-confident
due to excessive positive samples. So, we conducted research to address both the lack of labeled data and the false-positive increment issues to improve detection performance. This paper proposes a novel augmentation technique: \textbf{False-Positive (FP) Sampling}. FP sampling involves extracting samples of false positives from the model's predictions and incorporating them into subsequent training scenes. FP sampling can be considered a reasonable and intensive learning strategy as it leads the model to relearn examples on which it had previously made incorrect predictions. Fig. \ref{fig:Visualization} illustrates the results of data augmentation using GT and FP sampling methods.

In our experiments, we observed significant performance improvements in various models by applying FP
sampling. This finding underscores the effectiveness of FP sampling in reducing false positives in model
detection, independent of the underlying model architecture. This aspect of our research demonstrates the
adaptability and effectiveness of FP sampling in a broad range of model structures, highlighting its potential as a universal tool for enhancing detection accuracy in 3D object detection systems. Additionally, we verified the universal applicability of FP sampling across diverse datasets. FP sampling can achieve consistent performance despite each LiDAR dataset's unique format and class distribution and the wide variance in sensor performance.

The main contributions of this paper are summarized as follows:
\begin{itemize}
    \item We propose a new augmentation technique called FP sampling to reduce false positives in 3D object detection.
    \item We experimentally verify that FP sampling can be universally applied across various models and datasets.
\end{itemize}

\section{RELATED WORK}

\subsection{3D Object Detection}

In 3D object detection based from LiDAR data, various approaches have been developed with unique advantages. Point-based methods such as PointRCNN\cite{pointrcnn} and Part-A2\cite{part-a2} directly utilize raw 3D point cloud data. PointRCNN employs a two-stage architecture to refine initial 3D bounding box proposals and leverages the precise features of point clouds to achieve high accuracy. Part-A2 recognizes structural parts within objects to enhance detection performance. Voxel-based approaches like VoxelNet\cite{voxelnet}, SECOND\cite{second:}, and PointPillars\cite{pointpillars} transform point clouds into 3D grids and encode each voxel with volumetric representations. VoxelNet employs 2D convolutions for region proposal generation to extract meaningful features from voxelized data. SECOND is a network designed for efficient object detection, using voxel-based 3D convolutions for high-speed processing. PointPillars takes an innovative approach by dividing 3D space into vertical 'pillars,' simplifying the voxelization process and improving processing speed.

Unified approaches like PV-RCNN\cite{pvrcnn}, PV-RCNN++\cite{pvrcnnpp}, and CenterPoint\cite{centerpoint} combine the strengths of both point-based and voxel-based methods. PV-RCNN and PV-RCNN++ simultaneously process coarse-grained voxels and raw point clouds to maximize the advantages of both data representations. CenterPoint employs a keypoint detector to predict the geometric centers of objects, achieving more precise localization and improved speed.

In this paper, we selected PV-RCNN, SECOND, PointPillars, and CenterPoint to prove that our novel data augmentation and training approach overcomes the limitations of point cloud data and effectively improves the models object detection capabilities in diverse environmental conditions.

\subsection{Data Augmentation}

Data augmentation is a crucial strategy for improving the generalization ability of models and preventing the overfitting of neural networks. It has evolved across various fields. In 2D image processing, basic augmentation techniques such as flipping, rotation, and shifting are commonly used. At the same time, more complex methods like Copy-Paste\cite{RefWorks:RefID:11-ghiasi2021simple}, involving random copying and pasting of objects, have improved performance in image segmentation.

In contrast, research on augmenting 3D point cloud data is relatively in its early stages. Basic augmentation techniques for point cloud data include random scaling, flipping, rotation around the gravity axis, and point jittering. They have primarily been validated in 3D object detection \cite{hahner2020quantifying}. However, these basic augmentation techniques are limited in their ability to simulate complex real-world scenarios, requiring more sophisticated methods. In this context, methods such as PointWOLF \cite{RefWorks:RefID:13-kim2021point}, which involve performing locally weighted transformations at multiple anchor points to create smooth and realistic 3D point clouds, and a strategy using two cross-scan augmentations based on replacement and rotation to enhance LiDAR data effectively \cite{RefWorks:RefID:14-xiao2022polarmix:}, have been developed. 

Furthermore, a specialized approach for augmenting LiDAR point cloud data, Ground Truth Sampling, has been developed \cite{second:}. Studies like SECOND \cite{second:} have adopted this method for augmenting LiDAR data. Sampling actual object data and adding it to the training dataset is particularly effective in improving the generalization ability of models in complex 3D environments like autonomous vehicles. 

This paper proposes a new strategy beyond these augmentation methods: False-Positive Sampling. This new approach involves artificially creating points likely to be misrecognized and adding them to the dataset so that the model can better understand and respond to potential error situations that may occur in the real world.

\section{Method}

\begin{figure}[ht]
  \centering
  \includegraphics[width=0.5\textwidth]{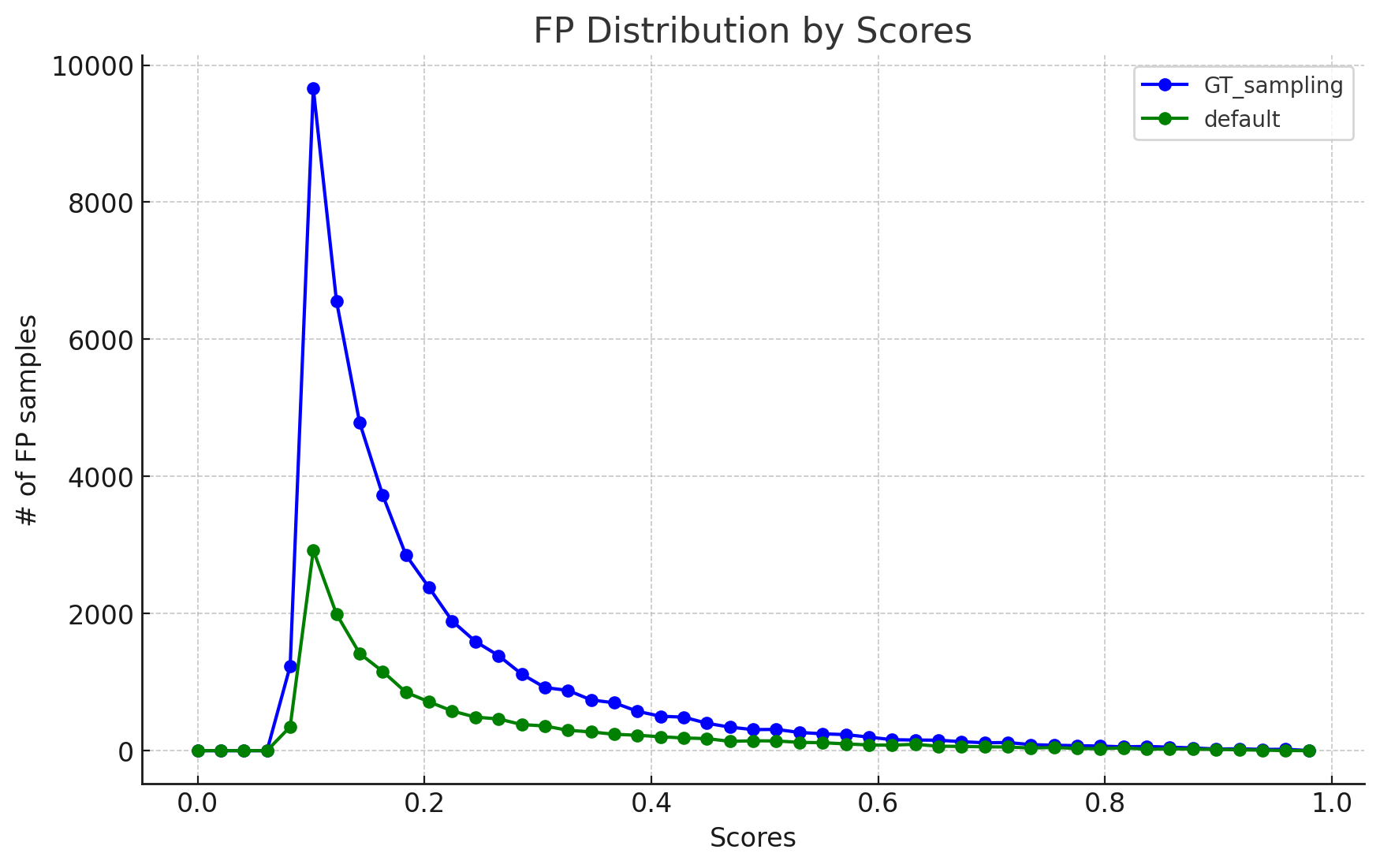}
  \caption{The distribution of false positives with or without GT sampling based on the model's confidence score. The model trained with GT sampling tends to be over-confident, if there's not a appropriate constraints to model's prediction.}
  \label{fig:fp_distribution}
\end{figure}

GT sampling\cite{second:} is a widely adopted data augmentation method in 3D object detection. This technique generates new training samples by replicating ground truth objects from the training dataset and repositioning them in various environments. As a result, it dramatically improves the model's ability to detect objects under diverse backgrounds and conditions. However, the model's excessive reliance on objects used in GT sampling can increase false positives, which are incorrect detections where the model erroneously identifies background or non-relevant objects as targets, potentially degrading the model's precision. As shown in Fig. \ref{fig:fp_distribution}, models employing GT sampling may experience more false positives than those without.

To solve this problem, we propose a new augmentation technique: \textbf{False Positive sampling (FP sampling)}. FP sampling is a strategy that actively utilizes the model's output in training. Samples that correspond to false positives in the model's predictions are stored in an FP database. These samples are then inserted into the point cloud of the scene used for training, enabling the model to learn that these samples are not objects to be detected. As a result of FP sampling, the trained model generates new false positive samples, which are subsequently stored in the FP database. By employing this process throughout training, in conjunction with GT sampling, we can effectively prevent the model from becoming overconfident in its predictions, thereby maintaining a balanced approach in detection tasks.

\begin{figure*}[ht]
  \centering
  \includegraphics[width=1.0\textwidth]{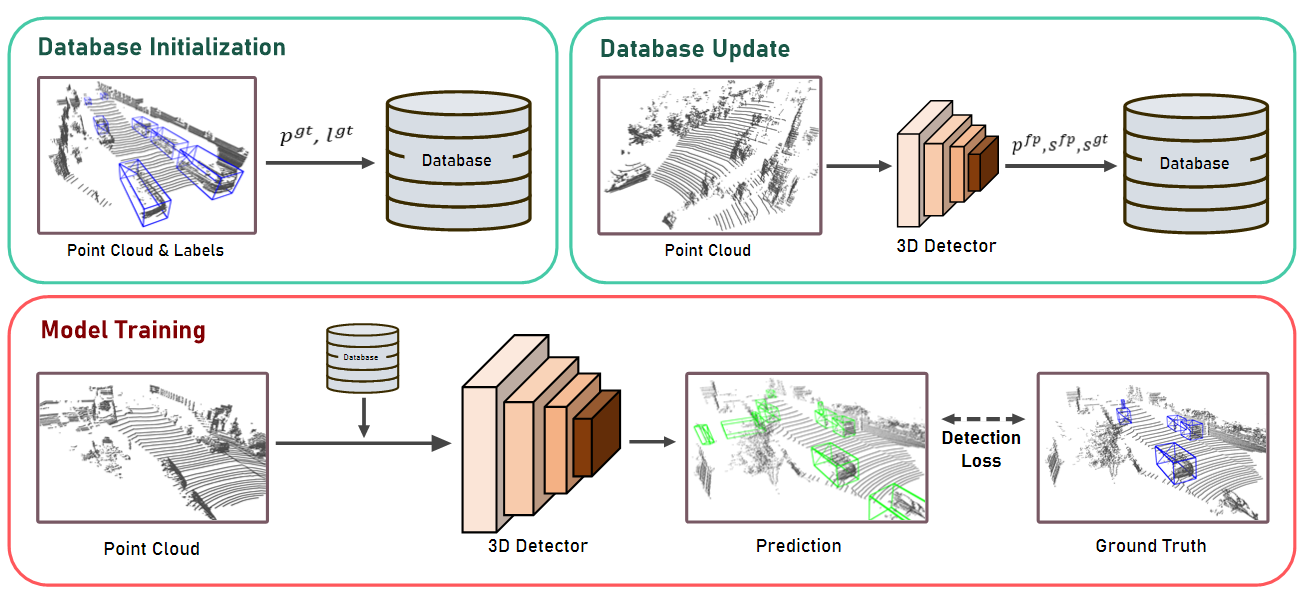}
  \caption{The proposed data augmentation and model training process, including the FP sample database management}
  \label{fig:framework}
\end{figure*}

This section explains specific procedures for performing FP sampling. As illustrated in Fig. \ref{fig:framework}, the FP sampling process includes three methods as follows:
\begin{itemize}
    \item Initialization and periodic update of the FP sample database during training runtime
    \item Insertion of FP sample point clouds into each scene
\end{itemize}

\subsection{Initialization and Update of FP Sample Database}
\label{sub:DB_update}

In this paper, an FP sample refers to an instance where the intersection over union (IoU) with the bounding box of every ground truth object in a given scene is zero, and an FP sample database means the storage of the FP samples. The basic method to construct the FP sample database is as follows: First, the model makes predictions for the entire training dataset. Next, the system finds the samples corresponding to false positives and stores them in the FP sample database. Note that the class of an FP sample is not an object's actual class but the model's predicted class.

FP sampling occurs at runtime to leverage the model's predictions. Even minor details in constructing the FP sample database can critically impact model performance. Therefore, several details must be observed when building the database to improve performance with FP sampling. FP sampling is a technique designed with the assumption that the model has achieved at least a certain level of performance in detecting objects. Therefore, the FP sample database is initialized only after sufficient training using GT sampling. After a required number of epochs, the FP sample database should be created for the first time through predictions on the entire dataset.

Suppose the initial FP sample database is continuously used throughout the training process. In that case, the model's generalization performance can be limited because it will only reflect the samples stored in the initial database. Therefore, it is necessary to update the database at some specific intervals. The database updating algorithm is described in Algorithm \ref{alg:fp_database_update}. To maximize the model's generalization ability, resetting and constructing the database can be more effective than just adding the model's new FP samples to the existing database.

\begin{algorithm}
\caption{Update of FP Sample Database}
\label{alg:fp_database_update}
\begin{algorithmic}
\State \textbf{Input:}
    \State $P = \{P_1,P_2,\ldots,P_i\}, \;P \in \mathbb{R}^3$ \hfill // original point cloud
    \State $f$: 3D object detector
\vspace{1mm}
    \State $\mathcal{D}^{gt}_c = \{(p^{gt}_1, l^{gt}_1), (p^{gt}_2, l^{gt}_2), \ldots], \;p^{gt}_i \in \mathbb{R}^3, \;l^{gt}_i \in \mathbb{R}^3$
\vspace{1mm}
    \State $\mathcal{D}^{gt} = \{\mathcal{D}^{gt}_1, \mathcal{D}^{gt}_2, \ldots, \mathcal{D}^{gt}_c\}$ \hfill // GT database
\vspace{1mm}
    \State $\mathcal{L}^{gt}_i = \{l'_1, l'_2, \ldots, l'_k\}$ \hfill // label set for each scene
\State \textbf{Output:}
\vspace{1mm}
    \State $\mathcal{D}^{fp}$: FP database
\end{algorithmic}

\begin{algorithmic}[1]
\Procedure{FP database update}{}
    \State $\mathcal{D}^{fp} \gets \varnothing$
    \For{$P_i \in P$}
        \State $Y \gets f(P_i), \;Y \in \mathbb{R}^3$
        \For{$y \in Y$}
            \If {$\textsc{IoUWithClosestGT}(y,\mathcal{L}^{gt}_i)=0$}
                \State $c \gets$ class of $y$
                \State add $p$ of $y$ to $\mathcal{D}^{fp}_c, \;p \in \mathbb{R}^3$ \hfill // p: FP   
            \EndIf
        \EndFor
    \EndFor
    \State \textbf{return} $\mathcal{D}^{fp}$
\EndProcedure
\end{algorithmic}
\end{algorithm}

\subsection{Insertion of Point Clouds of FP Samples} 
This section proposes a novel data augmentation algorithm that concurrently applies GT and FP sampling to each scene. This algorithm is designed to be used in conjunction with GT sampling. Initially, we utilize a GT sample database constructed before training and an FP sample database established during training, which will be discussed in \textit{Section} \textit{\ref{sub:DB_update}} in more detail. For a given point cloud $P$, we randomly select $\alpha_c$ samples from the GT database and $\beta_c$ samples from the FP database for each class $c_i \in C$ where $C=\{c_1,c_2,\ldots,c_n\}$. The selected samples are then concatenated with $P$ to form an augmented point cloud $P'$, and the $\alpha_c$ bounding boxes for the GT samples are integrated into the label set $L$. However, the bounding boxes for FP samples are not integrated into $L$ so that the model can learn that the point clouds of the FP samples do not represent actual objects. The data augmentation algorithm is described in Algorithm \ref{alg:gt_fp_sampling}.

\begin{algorithm}
\caption{Insertion of GT and FP Point Clouds}
\label{alg:gt_fp_sampling}
\begin{algorithmic}
\State \textbf{Input:}
    \State $P \in \mathbb{R}^3$: original point cloud
    \State $C = \{c_1, c_2, \ldots, c_n\}$ \hfill // set of classes
\vspace{1mm}
    \State $\mathcal{D}^{gt}_c = \{(p^{gt}_1, l^{gt}_1), (p^{gt}_2, l^{gt}_2), \ldots], \;p^{gt}_i \in \mathbb{R}^3, \;l^{gt}_i \in \mathbb{R}^3$
\vspace{1mm}
    \State $\mathcal{D}^{gt} = \{\mathcal{D}^{gt}_1, \mathcal{D}^{gt}_2, \ldots, \mathcal{D}^{gt}_c\}$ \hfill // GT database
\vspace{1mm}
    \State $\mathcal{D}^{fp}_c = \{p^{fp}_1, p^{fp}_2, \ldots\}, \;p^{gt}_i \in \mathbb{R}^3$
\vspace{1mm}
    \State $\mathcal{D}^{fp} = \{\mathcal{D}^{fp}_1, \mathcal{D}^{fp}_2, \ldots, \mathcal{D}^{fp}_c\}$ \hfill // FP database
    \State $\alpha_c$: number of GT samples for each class
    \State $\beta_c$: number of FP samples for each class
    \State $\mathcal{L} \in \mathbb{R}^3$: label for $P$
\State \textbf{Output:}
    \State $P' \in \mathbb{R}^3$: augmented point cloud
    \State $\mathcal{L}' \in \mathbb{R}^3$: label for $P'$
\end{algorithmic}

\begin{algorithmic}[1]
\Procedure{}{}
    \State $P' \gets P,\; \mathcal{L}' \gets \mathcal{L}$ 
    \For{$c \in C$}
        \For{$\alpha_c$}
            \State $p^{gt}, l^{gt} \gets \textsc{RandomSampling}(\mathcal{D}^{gt}_c)$
            \State $P' \gets P' \cup p^{gt},\; \mathcal{L}' \gets \mathcal{L}' \cup l^{gt}$
        \EndFor
        \For{$\beta_c$}
            \State $p^{fp} \gets \textsc{RandomSampling}(\mathcal{D}^{fp}_c)$
            \State $P' \gets P' \cup p^{fp}$
        \EndFor
    \EndFor
    \State \textbf{return} $P',\; \mathcal{L'}$
\EndProcedure
\end{algorithmic}
\end{algorithm}

\begin{figure*}[ht]
\centering
\subfloat[default]{\includegraphics[width=0.2\linewidth]{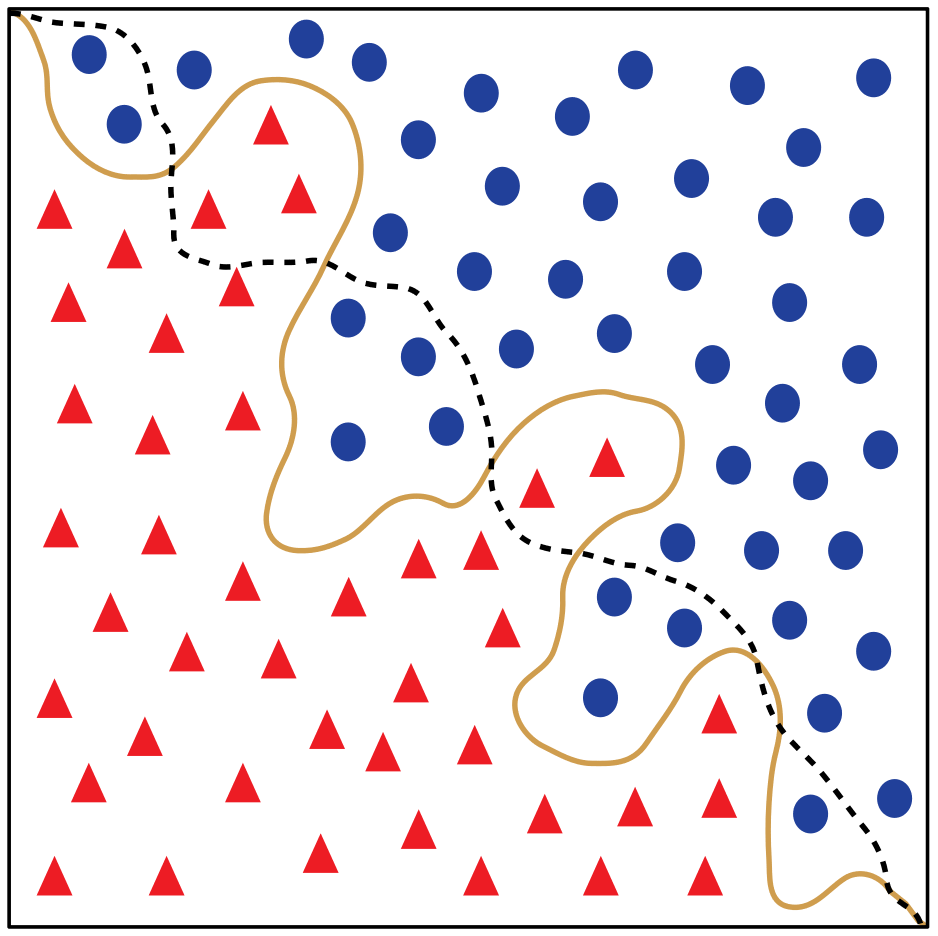}\label{fig:suba}}
\hfill
\subfloat[GT sampling]{\includegraphics[width=0.2\linewidth]{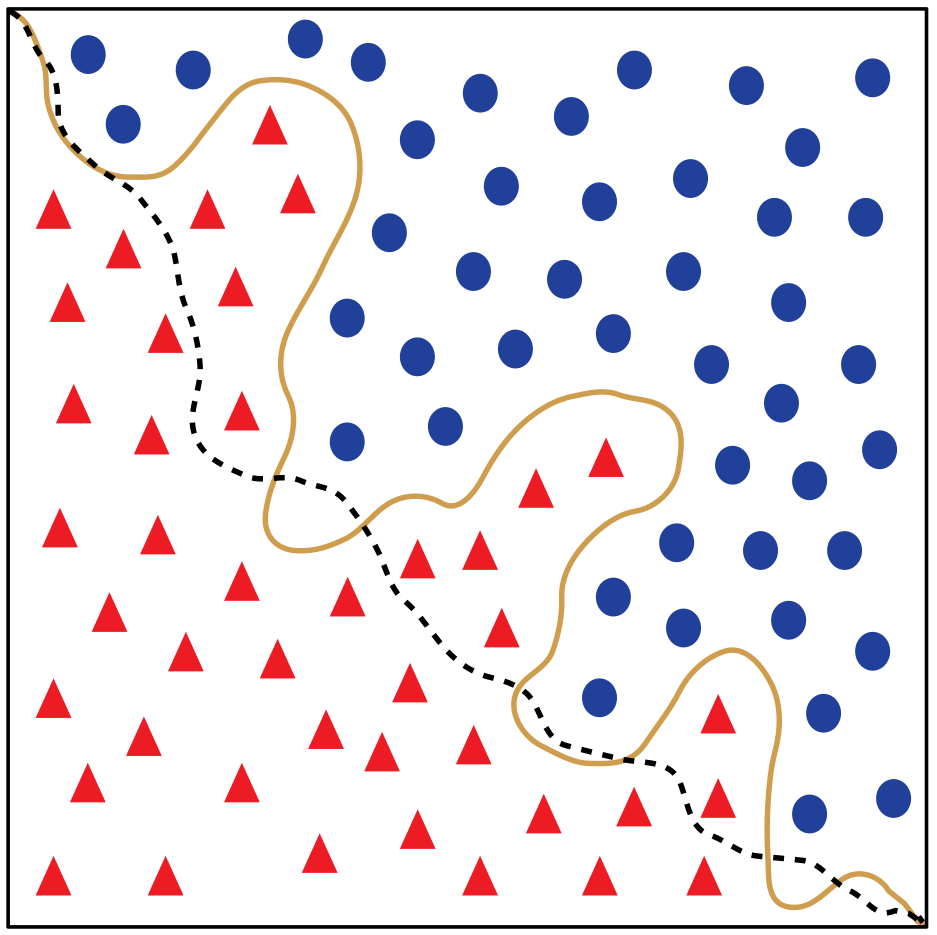}\label{fig:subb}}
\hfill
\subfloat[FP sampling]{\includegraphics[width=0.2\linewidth]{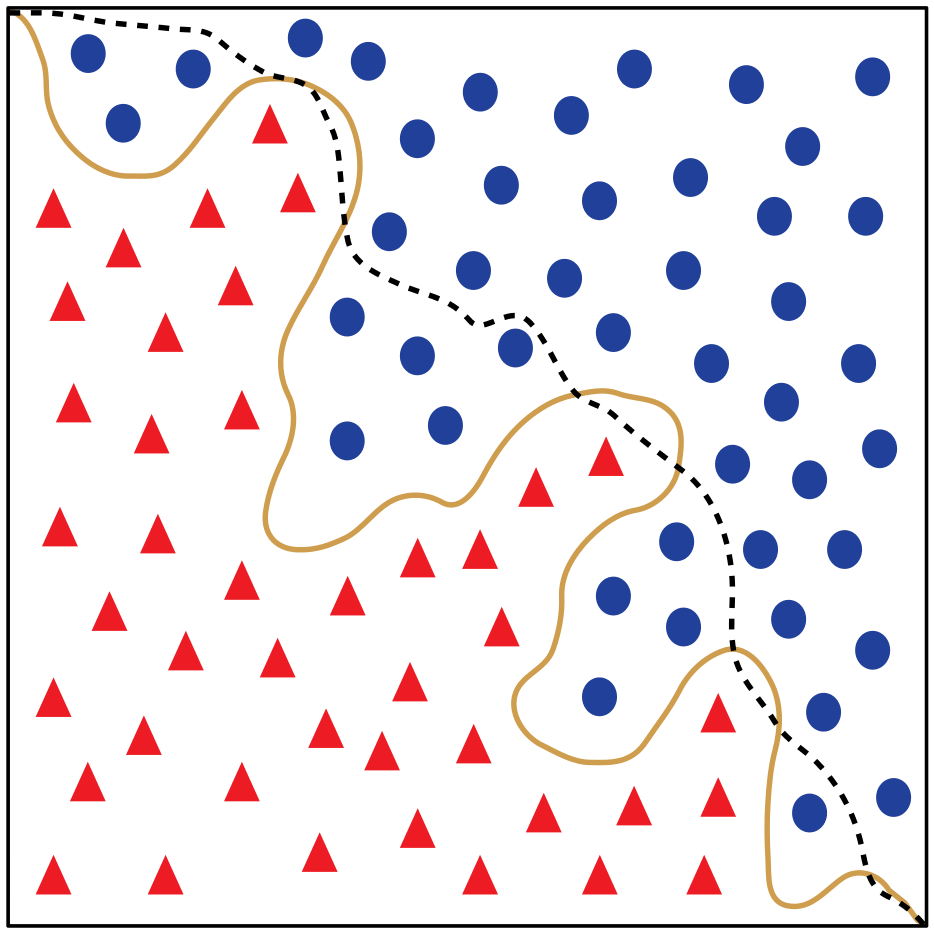}\label{fig:subc}}
\hfill
\subfloat[GT and FP sampling]{\includegraphics[width=0.2\linewidth]{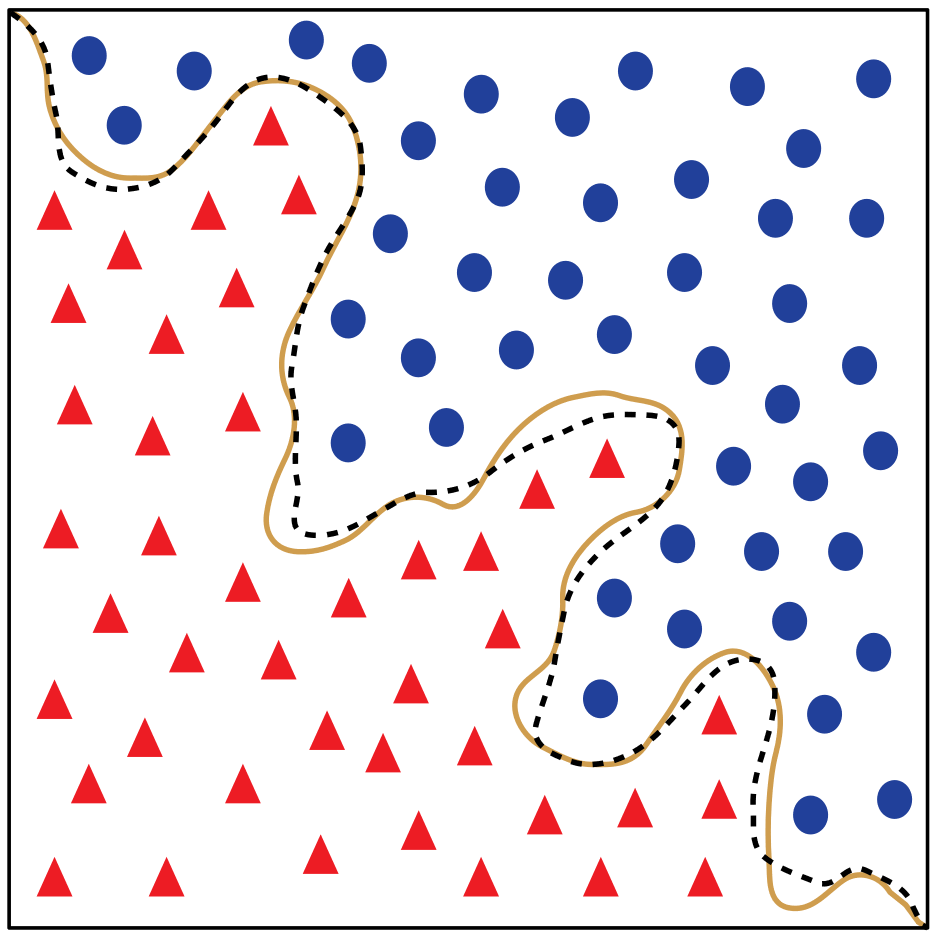}\label{fig:subd}}
\caption{The decision boundaries of the model changed by different data augmentation methods. Blue circles represent positive samples, while red triangles represent negative samples. Applying GT and FP sampling for data augmentation can modify the model's decision boundary (black dotted line) to be closer to the actual decision boundary (yellow solid line). Note that the figures result from dimensionality reduction from a very high dimensional space into two dimensions.}
\label{fig:decision_boundary}
\end{figure*}

Let us investigate how FP sampling can enhance detection performance in terms of the model's decision boundary. In the high-dimensional feature space for a point cloud, the decision boundary of the model exists to distinguish positive and negative samples. As illustrated in Fig. \ref{fig:decision_boundary}, the decision boundary of the model with GT sampling is biased towards increasing recall due to repetitive training on positive samples. Similarly, the model with FP sampling is biased towards increasing precision due to repetitive training on negative samples. Consequently, the model's decision boundary deviates from the actual decision boundary in either case. In contrast, the model trained with GT and FP sampling receives strong guidance to enhance both precision and recall, allowing for the refinement of weights to closely align the model's decision boundary with the actual decision boundary. Therefore, it tends to distinguish well between positive and negative samples.

\begin{figure*}[ht]
\centering
\subfloat[GT sampling]{\includegraphics[width=0.24\linewidth, height=0.25\linewidth]{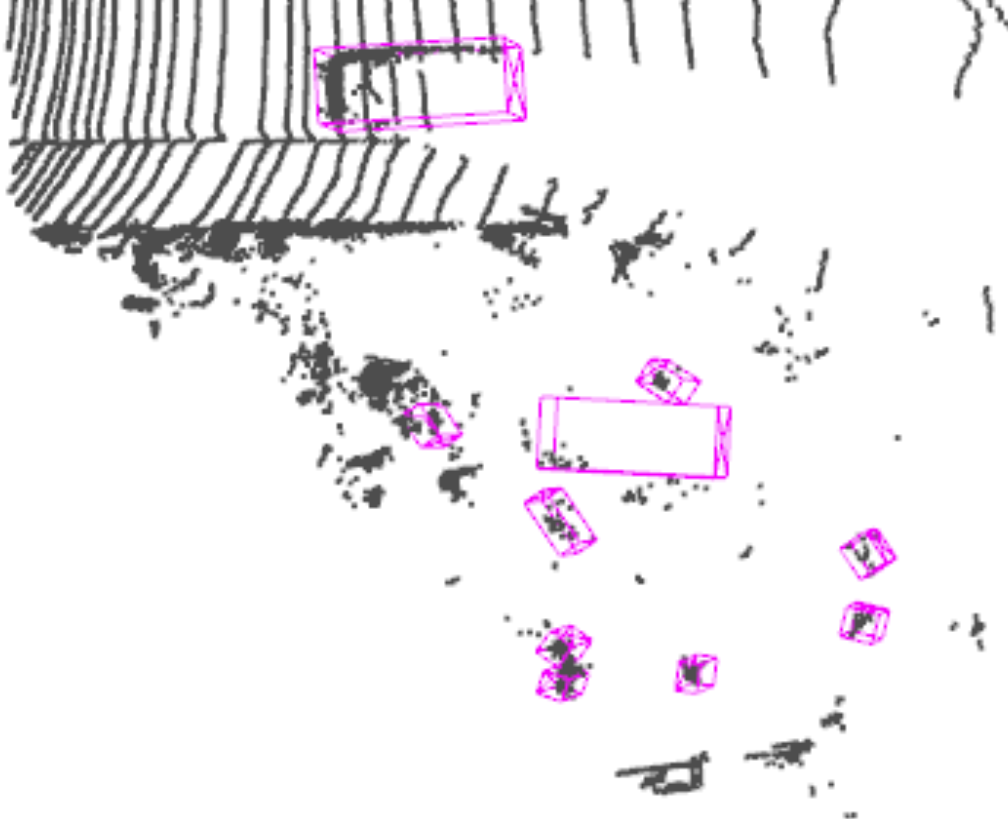}\label{fig:left1} \includegraphics[width=0.24\linewidth, height=0.25\linewidth]{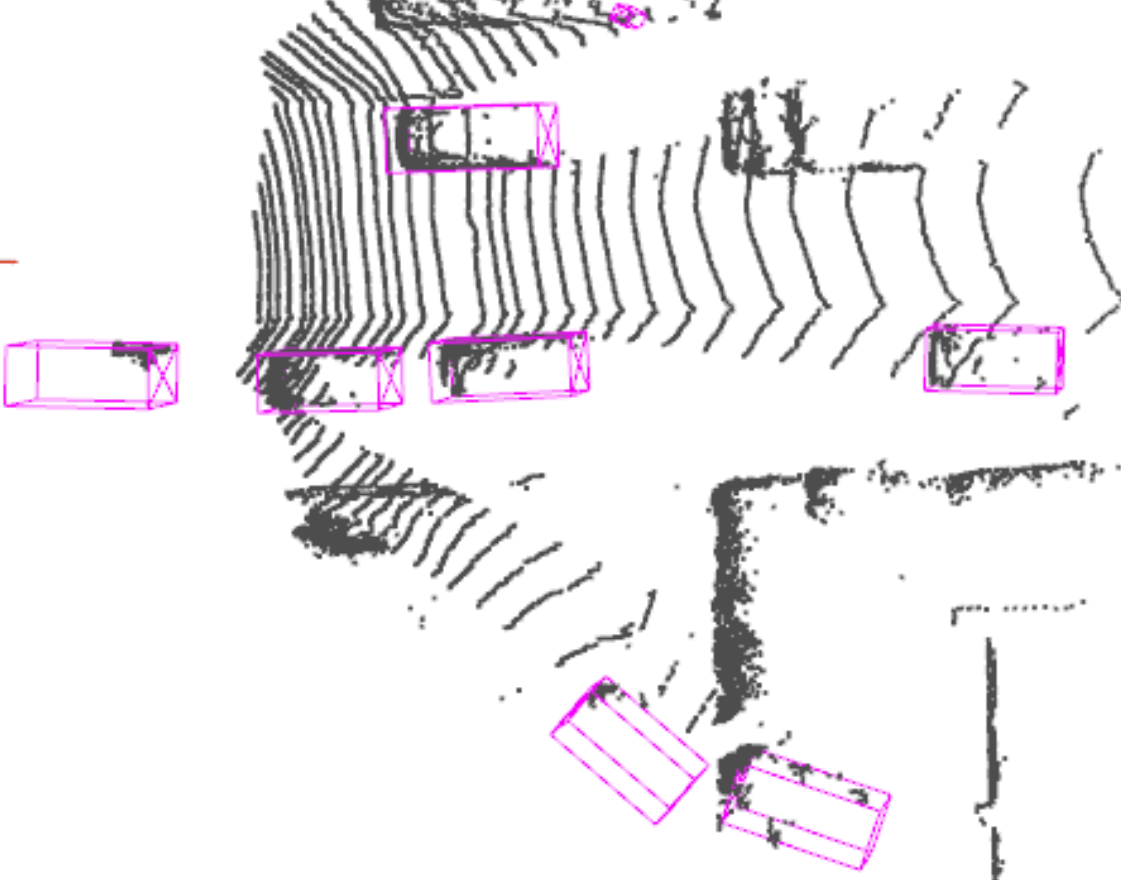}\label{fig:left2}}
\hfill
\subfloat[GT \& FP sampling]{\includegraphics[width=0.24\linewidth, height=0.25\linewidth]{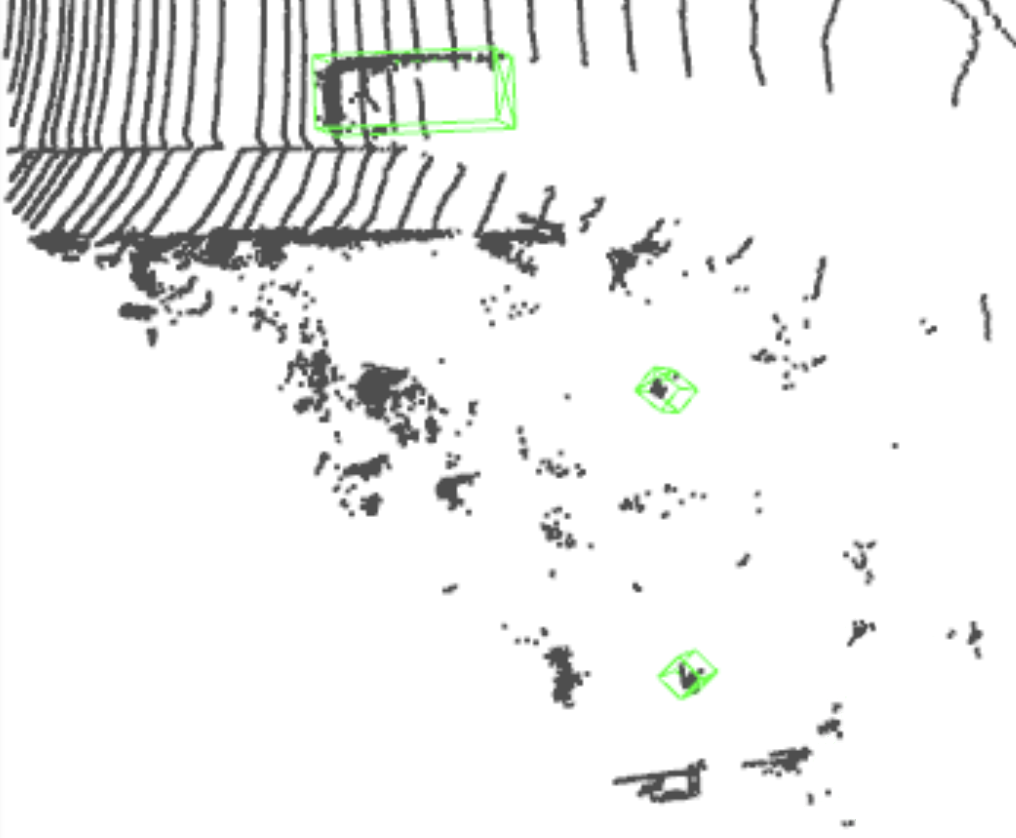}\label{fig:lefta} \includegraphics[width=0.24\linewidth, height=0.25\linewidth]{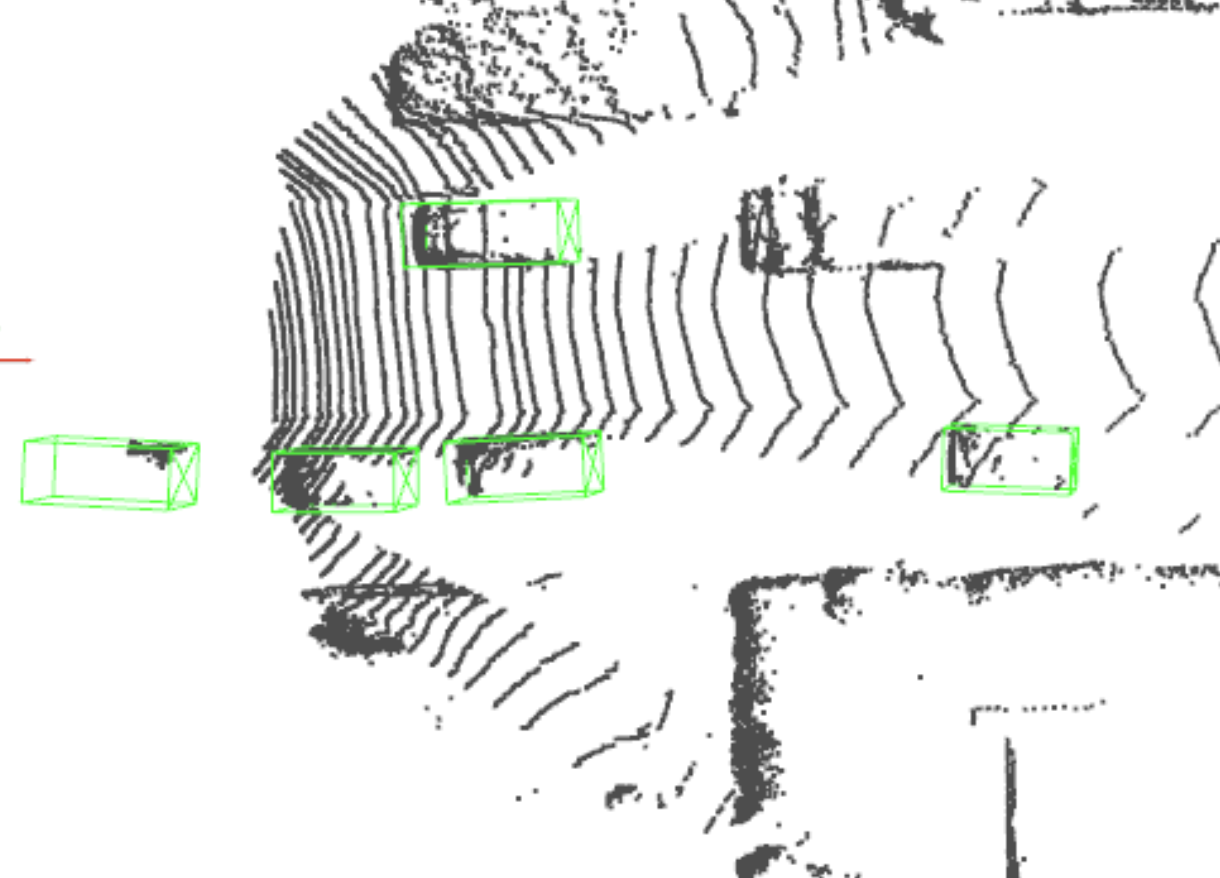}\label{fig:leftb}}
\caption{Comparison of GT sampling and GT \& FP sampling results on the KITTI validation dataset for the SECOND model. Pink and green bounding boxes represent the predictions made by the models trained with GT sampling and GT \& FP sampling methods, respectively.}
\label{visualize_point}
\end{figure*}

\section{EXPERIMENT}

\subsubsection{Implementation Details} 
In this study, we utilized an open-source repository, OpenPCDet\cite{openpcdet}, for 3D object detection. The methods implemented through OpenPCDet are designed to be easily applicable to various models. Specifically, for the KITTI dataset\cite{kitti}, we used the default training hyper-parameters of PointPillars\cite{pointpillars}, PV-RCNN\cite{pvrcnn}, SECOND\cite{second:}, and CenterPoint\cite{centerpoint} models, while for the Waymo dataset\cite{waymo}, we utilized the basic training hyper-parameters of the SECOND model. These models were trained for 80 epochs for KITTI dataset and 30 epochs for Waymo dataset, utilizing 4 NVIDIA GeForce RTX 4090 GPUs with a batch size of 8. During the training process, we employed the Adam Optimizer with a learning rate of 0.003, a weight decay of 0.01, and a momentum of 0.9.


\subsubsection{KITTI Dataset}
KITTI dataset\cite{kitti} used in this study comprises 7,481 training images and 7,518 test images, along with LiDAR point clouds, captured in urban and highway environments featuring a variety of road users such as vehicles, pedestrians, and cyclists. In the evaluation using this dataset, the mean Average Precision(mAP) of 3D bounding boxes was utilized as the primary evaluation metric. The mAP is an indicator of the accuracy of object detection models, measuring how precisely and accurately the models detect and classify objects. Particularly, this research assessed the models’ object detection performance within a 40-meter range using the mAP R40 metric. Using this metric is a crucial measurement method reflecting performance in real-road conditions akin to those in autonomous driving systems. Additionally, for the evaluation, mAP was further subdivided into three categories based on the difficulty level of each object category: easy, moderate, and hard. The IoU is a critical metric in assessing detection accuracy, indicating the overlap between the predicted object’s bounding box and the actual object’s bounding box. In this study, the IoU threshold was set to 0.7 for vehicles, and 0.5 for cyclists and pedestrians.

\subsubsection{Waymo Open Dataset}
Waymo Open Dataset\cite{waymo} was utilized to evaluate the 3D object detection capabilities critical to autonomous driving systems. This Dataset includes 798 training and 202 validation sequences, primarily providing data focused on vehicles and pedestrians. The dataset leverages a 64-ray LiDAR system, generating approximately 180,000 LiDAR points every 0.1 seconds, allowing for a detailed assessment of object detection performance under various road conditions and environments. The 3D detection evaluation metrics include the standard mAP for 3D bounding boxes and the mAP weighted by heading accuracy (mAPH). These metrics evaluate vehicles with an IoU threshold of 0.7 and pedestrians with an IoU threshold of 0.5. The Waymo dataset offers two levels of difficulty for performance analysis, categorized as LEVEL 1 and LEVEL 2. LEVEL 1 targets boxes with more than five LiDAR points, while LEVEL 2 includes boxes with at least one LiDAR point. Due to the vast data in the dataset, this research utilized only about 20\% of the training data for experiments. This selection was made to increase the efficiency of the research, focusing on securing a sufficient amount of data to evaluate the model`s performance adequately.
\begin{table*}[ht]
\caption{Comparison of performance for Car, Pedestrian, and Cyclist classifications in various 3D object detection models based on the application of FP sampling on KITTI dataset. This table utilizes SECOND, PV-RCNN, PointPillars, and CenterPoint models, depicting the mAP when FP sampling is applied and not applied, detailed across the R40 benchmarks: Easy, Moderate, and Hard.}
\centering
\renewcommand{\arraystretch}{1.3} 
\begin{tabular}{cccccccccc}
\hline
\multirow{2}{*}{\textbf{Method}}    & \multicolumn{3}{c}{\textbf{Car - 3D Detection}}                        & \multicolumn{3}{c}{\textbf{Pedestrian - 3D Detection}}                 & \multicolumn{3}{c}{\textbf{Cyclist - 3D Detection}}                    \\ \cline{2-10} 
                                    & \textbf{Easy}    & \textbf{Mod.} & \textbf{Hard}    & \textbf{Easy}    & \textbf{Mod.} & \textbf{Hard}    & \textbf{Easy}    & \textbf{Mod.} & \textbf{Hard}    \\ \hline
Pointpillars\cite{pointpillars} + GT sampling          & \textbf{89.99}   & 78.60    & 75.80   & 55.69            & 49.22             & 44.86            & 80.80            & 61.99             & 57.21            \\
Pointpillars + GT \& FP sampling (Ours) & 87.95            & \textbf{78.79}             & \textbf{76.22}            & \textbf{56.91}   & \textbf{52.08}    & \textbf{47.62}   & \textbf{86.22}   & \textbf{67.17}    & \textbf{62.64}   \\
\rowcolor{skyblue}
\textit{Improvement}                         & \textit{-2.04} & \textbf{\textit{+0.19}}  & \textbf{\textit{+0.42}} & \textit{\textbf{+1.22}} & \textit{\textbf{+2.86}}  & \textit{\textbf{+2.76}} & \textit{\textbf{+5.42}} & \textit{\textbf{+5.18}}  & \textit{\textbf{+5.43}} \\ \hline
Centerpoint\cite{centerpoint} + GT sampling           & 88.48   & 79.82             & 77.41            & 53.65            & 49.73             & 46.14            & 79.86            & 64.00    & \textbf{60.98}   \\
Centerpoint + GT \& FP sampling (Ours)  & \textbf{89.54}            & \textbf{81.04}    & \textbf{78.73}   & \textbf{62.87}   & \textbf{58.78}    & \textbf{53.95}   & \textbf{80.48}   & \textbf{64.75}            & 60.22           \\
\rowcolor{skyblue}
\textit{Improvement }                        & \textit{\textbf{+1.06}} & \textbf{\textit{+1.22}}  & \textbf{\textit{+1.32}} & \textbf{\textit{+9.22}} & \textbf{\textit{+9.05}}  & \textbf{\textit{+7.81}} & \textbf{\textit{+0.62}} & \textit{\textbf{+0.75}}  & \textit{-0.76} \\ \hline
SECOND\cite{second:} + GT sampling                & 89.96            & 81.16             & 78.39            & 54.37            & 49.44             & 45.25            & 82.06            & 65.62             & 61.63            \\
SECOND + GT \& FP sampling (Ours)       & \textbf{90.96}   & \textbf{82.16}    & \textbf{79.41}   & \textbf{61.61}   & \textbf{57.22}    & \textbf{51.64}   & \textbf{88.15}   & \textbf{67.31}    & \textbf{63.08}   \\
\rowcolor{skyblue}
\textit{Improvement }                        & \textbf{\textit{+1.00}} & \textbf{\textit{+1.00}}  & \textbf{\textit{+1.02}} & \textbf{\textit{+7.24}} & \textbf{\textit{+7.78}}  & \textbf{\textit{+6.39}} & \textbf{\textit{+6.09}} & \textbf{\textit{+1.69}}  & \textbf{\textit{+1.45}} \\ \hline
PV-RCNN\cite{pvrcnn} + GT sampling               & \textbf{92.00}   & 84.38             & 82.38            & 68.81            & 56.72             & 51.63            & \textbf{90.86}   & \textbf{72.55}    & \textbf{68.05}   \\
PV-RCNN + GT \& FP sampling (Ours)      & 91.94            & \textbf{84.86}    & \textbf{82.65}   & \textbf{69.99}   & \textbf{61.53}    & \textbf{56.64}   & 89.27            & 70.25             & 65.91            \\
\rowcolor{skyblue}
\textit{Improvement }                        & \textit{-0.06} & \textbf{\textit{+0.48}}  & \textbf{\textit{+0.27}} & \textbf{\textit{+1.18}} & \textbf{\textit{+4.81}}  & \textbf{\textit{+5.01}} & \textit{-1.59} & \textit{-2.30}  & \textit{-2.14} \\ \hline
\end{tabular}
\label{table:table1}
\end{table*}

\begin{table*}[ht]
\caption{Comparison of vehicle, pedestrian, and cyclist detection performance using the SECOND models with and without FP sampling on the Waymo dataset. This table compares vehicle, pedestrian, and cyclist detection performance in the Waymo dataset using the SECOND model with and without the application of FP sampling. The effectiveness of FP sampling is assessed using the mAP at two levels of difficulty: Level 1(L1) and Level 2(L2).}
\centering
\renewcommand{\arraystretch}{1.3} 
\begin{tabular}{ccccccc}
\hline
\multirow{2}{*}{\textbf{Method}}                             & \multicolumn{2}{c}{\textbf{Vehicle - 3D Detection}}  & \multicolumn{2}{c}{\textbf{Pedestrian - 3D Detection}} & \multicolumn{2}{c}{\textbf{Cyclist - 3D Detection}}  \\ \cline{2-7} 
                                                     &\textbf{Vec\_L1}              & \textbf{Vec\_L2}              & \textbf{Ped\_L1}               & \textbf{Ped\_L2}               & \textbf{Cyc\_L1}              & \textbf{Cyc\_L2}              \\ \hline
SECOND\cite{second:} + GT sampling                   & 70.07                & 61.86                & 64.85                 & 57.01                 & 56.74                & 54.63                \\
SECOND + GT \& FP sampling (Ours)                        & \textbf{70.94}                & \textbf{62.66}                & \textbf{65.58}        & \textbf{57.63}        & \textbf{57.78}       & \textbf{55.61}       \\
\rowcolor{skyblue}
\textit{Improvement}                                 &  \textit{\textbf{+0.87}}                    &      \textit{\textbf{+0.80} }               &      \textit{\textbf{+0.73}}                 &      \textit{\textbf{+0.62}}                 &        \textit{\textbf{+1.04}}              &     \textit{\textbf{+0.98}}                \\ \hline
Pointpillars\cite{pointpillars} + GT sampling                   & 70.11                & 61.93                & 65.72                 & 57.60                 & 53.35                & 51.32                \\
Pointpillars + GT \& FP sampling (Ours)                        & \textbf{70.55}                & \textbf{62.32}                & \textbf{67.60}        & \textbf{59.33}        & \textbf{59.33}       & \textbf{57.10}       \\
\rowcolor{skyblue}
\textit{Improvement}                                 &  \textit{\textbf{+0.44}}                    &      \textit{\textbf{+0.29} }               &      \textit{\textbf{+1.88}}                 &      \textit{\textbf{+1.67}}                 &        \textit{\textbf{+5.98}}              &     \textit{\textbf{+5.78}}                \\ \hline
\end{tabular}
\label{table:table2}
\end{table*}

\subsection{Result Analysis}

In this study, we analyzed the performance of 3D object detection targeting cars, pedestrians, and cyclists using the KITTI dataset. Specifically, we compared the GT sampling and GT \& FP sampling methods across four prominent detection models: PointPillars, CenterPoint, SECOND, and PV-RCNN. The analysis focused on the moderate difficulty level of each class, reflecting a balanced scenario in practical applications. The results of this analysis can be found in Table \ref{table:table1}. A notable finding from this study is that most models showed improved performance with the GT \& FP sampling method compared to GT sampling. This improvement was particularly evident in the SECOND model, as illustrated in Fig. \ref{visualize_point}. In this model, the car class recorded 82.16\%, the pedestrian class 57.22\%, and the cyclist class 67.31\%, indicating performance enhancements across all difficulty levels and classes. These results suggest that the GT \& FP sampling method plays a significant role in improving the performance of object detection models.

Furthermore, a distinctive feature observed in the study results is the substantial improvement in the pedestrian class across all models. For example, the Pointpillars model showed an increase from 49.22\% with GT sampling to 52.08\% with GT \& FP sampling. A similar trend was observed in the CenterPoint model, where performance rose from 49.73\% to 58.78\% by applying GT \& FP sampling. The primary reason for this notable enhancement in pedestrian detection is the tendency of pedestrian objects to exhibit a high rate of false positives in the early stages of training. The GT \& FP sampling method significantly reduced these false positives, leading to the observed performance improvements in the pedestrian category. This finding suggests that reducing false positives during the training process of object detection models plays a critical role in enhancing performance, providing important guidelines for future improvements in object detection models.

In evaluating performance on the Waymo dataset, the versatility of this augmentation is substantiated, demonstrating its efficacy with the utilization of SECOND models in 3D object detection. Table \ref{table:table2} presents each class`s GT and GT \& FP sampling performance, encompassing vehicles, pedestrians, and cyclists. Considering the Level 2 (L2) metric, applying GT \& FP sampling in the SECOND model exhibited an mAP improvement of 0.24\% for vehicles, 0.65\% for pedestrians, and 1.43\% for cyclists. These results suggest that such methods are effective in datasets like KITTI and can be efficiently applied in large-scale datasets.

In conclusion, the FP sampling technique proposed in this study has played a significant role in enhancing the performance of 3D object detection models. Our experimental evaluations on the KITTI and Waymo datasets have demonstrated that this method can improve performance in various models, with an exceptionally remarkable enhancement in pedestrian detection. However, it is essential to note that not all object classes exhibited performance improvements. This variation might be due to factors such as the differences in the distribution of false positives and true positives across models, and the variability in the number of samples and characteristics among classes. These aspects highlight challenges in hyperparameter tuning and suggest areas for further research and refinement.

\subsection{Ablation Studies}
In this section, we perform a performance analysis of 3D object detection models using the KITTI dataset\cite{kitti}. In particular, we investigate the impact of FP sampling on the capability of 3D object detection in autonomous driving scenarios, utilizing the SECOND \cite{second:}. All experiments are conducted using a consistent dataset and model, ensuring the coherence and comparability of the research. This study aims to quantitatively assess the influence of FP sampling strategies on the accuracy and overall performance of 3D object detection.\\

\subsubsection {Impact of FP Sampling Strategies$\colon$Impact of FP sampling on False Positive Reduction} 

\begin{figure}[ht]
  \centering
  \includegraphics[width=0.5\textwidth]{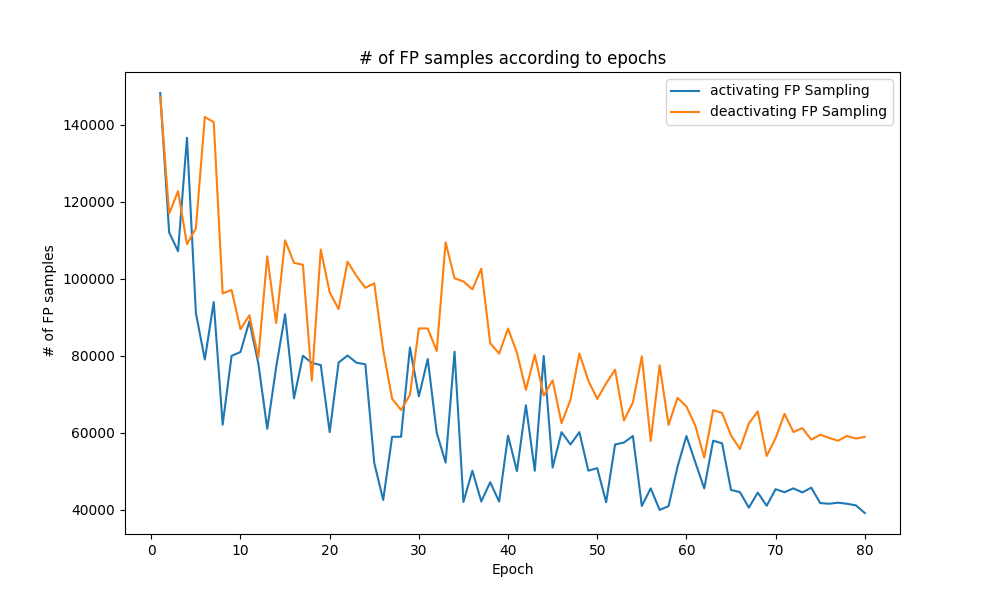}
  \caption{Variation in the number of FP samples over epochs based on the application of FP sampling. This graph's horizontal axis represents epochs, while the vertical axis indicates the number of FP samples. The blue line depicts the scenario with FP sampling applied, whereas the orange line represents the scenario without FP sampling.}
  \label{fig:fp_by_epoch}
\end{figure}

In this section, we analyze the impact of False Positive (FP) sampling on the number of false positives generated by the model. The main results of this analysis can be observed in Fig. \ref{fig:fp_by_epoch}, which visually demonstrates the role of FP sampling in effectively reducing the number of false positives in the model. The graph compares scenarios with and without applying FP sampling, showing the change in the number of false positives over epochs. A significant reduction in false positives is observed in the model where FP sampling is applied. This indicates that FP sampling is an effective technical strategy to reduce false positives in the model's learning process.

\subsubsection{Performance Enhancement through FP Sampling Integration\:} 

Following the confirmation in the previous section that FP sampling effectively reduces the number of false positives, we now focus on how this reduction impacts the overall performance improvement of the model. In this section, we conduct a detailed analysis based on the data presented in Table \ref{table:table3}, which compares the performance of the model with neither GT nor FP sampling applied, only GT sampling applied, only FP sampling applied, and both GT \& FP sampling applied.

The analysis shows that applying FP sampling independently does not significantly enhance the model's
performance. This finding suggests that while FP sampling effectively reduces false positives, it overlooks other important elements when used independently. However, when GT sampling and FP sampling were combined, the advantages of both methods complemented each other, leading to an overall improvement in model performance.

These results indicate that a harmonious combination of FP sampling and GT sampling can play a crucial role in enhancing the accuracy and reliability of 3D object detection models. This observation demonstrates that FP sampling, when used with GT sampling, can overcome the limitations of being used alone and effectively improve the model's detection capabilities. The analysis of Table III confirms that the appropriate combination of FP sampling and GT sampling is an effective approach in 3D object detection. This finding provides new guidelines for utilizing FP sampling in future developments of 3D object detection models.

\begin{table}[ht]
\caption{Performance of 3D object detection models with FP sampling, GT sampling, and GT \& FP sampling. This table presents the mAPs for the car, pedestrian, and cyclist classes, focusing specifically on the moderate difficulty level.}
\centering
\begin{tabular}{cc|ccc} 
\hline
GT sampling               & FP sampling                & Car   & Pedestrian & Cyclist \\ \hline 
            &                      & 75.02  & 46.33       & 48.67    \\
                          &\usym{2713}                          & 78.31 & 46.53      & 48.39   \\
\usym{2713}                          &                            & 82.01 & 53.14      & 64.26   \\
\usym{2713}                          &\usym{2713}                            & \textbf{82.03} & \textbf{55.62}     & \textbf{65.31}   \\ \hline
\end{tabular}
\label{table:table3}
\end{table}

\section{CONCLUSION}
This study demonstrates that the FP sampling improves the performance of 3D object detection significantly. FP sampling contributes to performance improvement by reducing false positives and enhancing data augmentation. This strategy is expected to offer practical applications in areas such as autonomous driving, robotics, and surveillance systems and to significantly impact the advancement of computer vision and artificial intelligence \cite{Lee:2020StrDAN, Lee:2023JCDE_MUSP, Lee:2020JCDE, Eom:2022JCDE}. However, this approach presents two main limitations. Firstly, the requirement to update the database at specific epochs leads to significant consumption of computational resources, including time cost. Secondly, including FP sampling techniques increases the number of hyperparameters specifically used for sampling, thereby complicating the process. These limitations are pivotal for future research to enhance model efficiency and effectiveness.

\bibliographystyle{ieeetr}
\bibliography{export}

\addtolength{\textheight}{-12cm}   

\end{document}